\documentclass[letterpaper, 10 pt, conference]{ieeeconf}  

\IEEEoverridecommandlockouts                              

\newcommand{\eg}{\textit{e.g.}}

\usepackage{graphicx}
\usepackage{subcaption}
\usepackage{multirow}
\usepackage{xcolor}
\usepackage{amsmath}
\usepackage{hyperref}

\title{\LARGE \bf
SCOUT: Semantic scene COverage via Uncertainty-guided Traversal
}

\author{Junyu Mao$^{1,*,\dagger}$, Sara Ayoubi$^{1,*}$, Vishnu D. Sharma$^{2}$, Ilija Hadžić$^{2,\ddagger}$, and Matthew Andrews$^{2}$%
\thanks{This work was presented at the 2026 ICRA Workshop on Uncertainty in Open World Robotics.}%
\thanks{$^{*}$Equal contribution.}%
\thanks{$^{\dagger}$Work done during an internship at Nokia Bell Labs France. Junyu Mao is also with the Department of Electrical and Electronic Engineering, Imperial College London, London, UK.}%
\thanks{$^{\ddagger}$ Ilija Hadžić is currently with Locus Robotics. The work described in this paper was performed in its entirety while he was affiliated with Nokia Bell Labs, USA.}%
\thanks{$^{1}$Nokia Bell Labs, France.}%
\thanks{$^{2}$Nokia Bell Labs, Murray Hill, NJ, USA.}%
\thanks{Supplementary video: \protect\url{https://bit.ly/4mJgI5T}}%
}

\begin{document}

\maketitle
\thispagestyle{empty}
\pagestyle{empty}

\begin{abstract}
Robots that operate over extended periods should not merely visit space; they should progressively understand it. Yet most 3D scene graph pipelines treat perception as a post-processing stage over a fixed dataset, decoupling scene representation from the decisions that determine what is observed in the first place. We present SCOUT, an online semantic exploration framework that closes this loop by coupling active traversal with probabilistic scene graph construction. Given a prior 2D occupancy map and posed RGB-D observations, SCOUT incrementally builds an uncertainty-aware 3D scene graph whose nodes maintain fused geometry and posterior beliefs over open-vocabulary object labels, while edges encode structural relations such as \textit{on}, \textit{inside}, \textit{belong}, and \textit{next to}. These beliefs are fed back to an uncertainty-guided traversal planner, which selects viewpoints by balancing expected semantic certainty gain, geometric coverage gain, and travel cost. In this way, the robot revisits ambiguous objects when additional evidence matters and expands into unseen free space when the scene remains incomplete. The resulting system treats semantic scene completeness as an operational objective rather than a passive by-product of semantic mapping, moving toward autonomous agents that can patrol, update, and reason about evolving indoor environments with minimal human intervention.
\end{abstract}

\section {Introduction}

It is of great significance for robots to understand the environments (or scenes) in which they operate, as this empowers them to perform
complex tasks of reasoning, querying, and interacting with objects. To this end, semantic scene representations have been extensively studied to
encode both the geometric (\eg~the shapes, locations, and spatial relationships of objects) and semantic information (\eg~the meanings, categories, and
functional relationships of objects) of the environment. Such comprehensive representations facilitate complex, long-term, open-vocabulary tasks,
potentially under high-level natural language commands from humans. Among them, 3D scene graphs \cite{armeni20193d} have been widely adopted as a
compact and expressive representation for robot reasoning.

Most existing research has focused on constructing and synthesizing static representations of scenes from sensory observations \cite{hughes2022hydra,
gu2024conceptgraphs, werby2024hierarchical, linok2024beyond}.
However, the common approach adopted by most scene-generation literature is that the scene-graph generation pipeline almost always follows the same
pattern: they assume the data is collected once (either manually or via an existing dataset), and then process that data to generate the scene graph
using existing computer-vision tools for object detection, segmentation, and labeling. As a result, scene understanding is decoupled from the process
that determines what is observed in the first place.

In this paper, we address the problem of integrating data collection directly into the scene-graph generation pipeline. Our motivation is twofold: (1) to enable autonomous recognition of previously unseen scenes without a human in the loop, which is critical when manual inspection is costly or the environments are dangerous; and (2) to enable continual refinement of the scene graph through successive observations, allowing object semantics, geometry, and spatial relations to be updated as the robot gathers additional evidence. Moreover, this coupled approach can mitigate the inefficiency of collecting data first, identifying blind spots only after scene-graph generation, and subsequently requiring additional data collection to fill those gaps.

We assume in our problem setting that robots already have a 2D map of the space, and the goal is to generate a semantic understanding of the objects
in the environment and the spatial relationships between them. While one might assume that an exhaustive traversal strategy would suffice to scan the
space and produce the necessary inputs for the scene-graph generation pipeline, such traversal can still be inefficient and insufficient to obtain a complete
semantic understanding. In particular, occlusions may require observing objects from one or multiple viewpoints, while exhaustive coverage can be time-consuming
and produce redundant captures in regions that contain no objects. 

To this end, we propose SCOUT, a semantic-coverage driven
traversal approach. SCOUT maintains an uncertainty-aware 3D scene graph through a probabilistic scene graph generator (PSGG), and uses an uncertainty-guided
traversal (UGT) planner to select viewpoints by balancing expected certainty gain, geometric coverage gain, and travel cost. In this way, observations are processed in real time, and the outcome of this processing drives which parts of the scene to explore next. By doing so, SCOUT is also connected to active learning, where an agent selects informative samples under a limited budget \cite{settles2009active}; here, each queried sample is an embodied viewpoint whose value depends on the considered factors. We introduce semantic scene completeness as a guiding principle for
data collection, enabling the robot to revisit ambiguous objects when additional evidence is valuable while continuing to expand into unseen free space. 

\begin{figure*}[!t]
\centering
\includegraphics[width=0.9\textwidth]{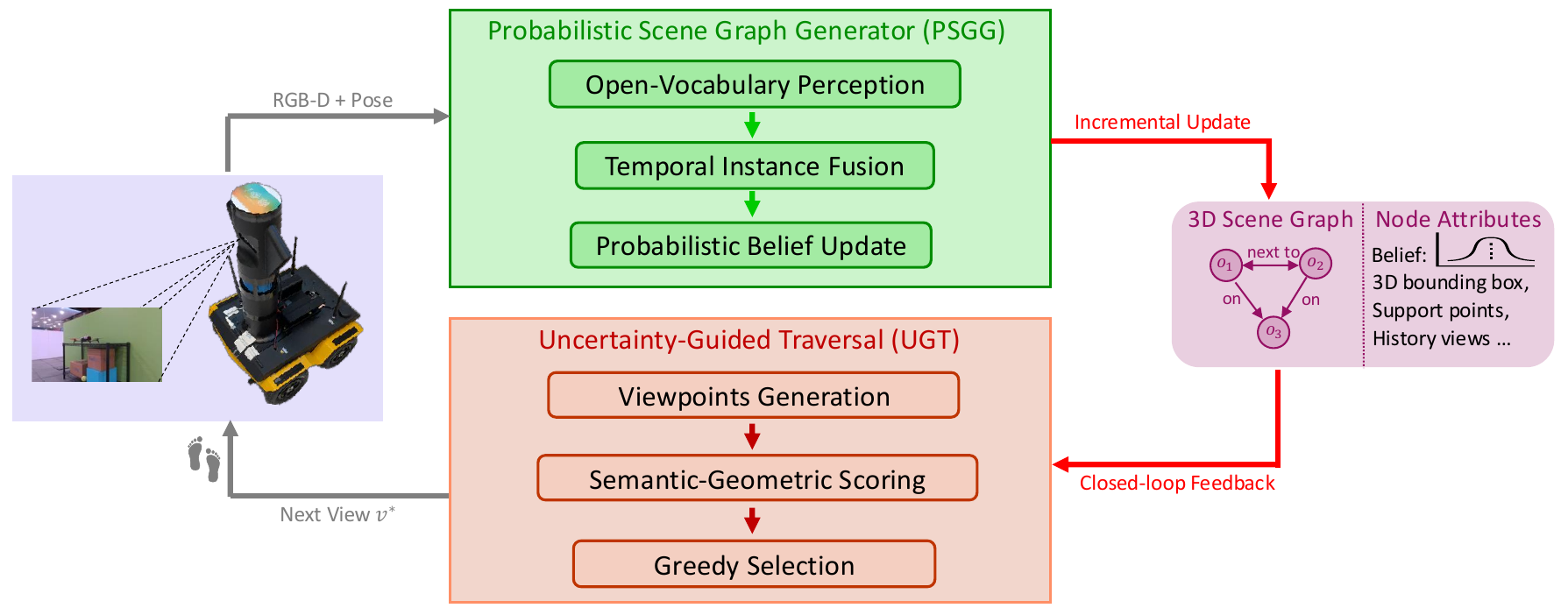}
\caption{System architecture of SCOUT. At each step, posed RGB-D observations are processed by PSGG through open-vocabulary perception, temporal instance fusion, and probabilistic semantic belief update to incrementally maintain an uncertainty-aware 3D scene graph. The graph stores object-level semantic beliefs, fused 3D geometry, spatial relations, and view histories. UGT consumes this graph to generate candidate viewpoints and evaluate them using expected certainty gain, geometric coverage gain, and travel cost, after which the highest-scoring viewpoint is selected as the next view. The resulting observation is fed back to PSGG, forming a closed active-perception loop.}
\label{fig:system-arch}
\end{figure*}



\section{Method}
SCOUT enables continual refinement of scene representations to maximize scene understanding completeness. It comprises two coupled components: an uncertainty-guided traversal (UGT) planner that selects viewpoints by balancing semantic exploitation and geometric exploration, and a probabilistic scene graph generator (PSGG) that incrementally integrates posed RGB-D observations into an uncertainty-aware 3D scene graph. Rather than deterministic estimates, the graph maintains a probabilistic scene belief for semantic querying, navigation, and manipulation, enabling active perception with incremental reasoning (Fig.~\ref{fig:system-arch}). 

We first provide a high-level overview of SCOUT and then describe each component in detail. SCOUT is an iterative system that alternates between the UGT (\S~\ref{sec:UGT}) and PSGG (\S~\ref{sec:PSGG}) modules. The UGT determines the next viewpoint to visit, while the PSGG constructs and maintains the scene graph. Each iteration begins with the UGT selecting the next viewpoint (\S~\ref{subsec:viewpoint_generation}). Initially, the candidate set contains only geometrically generated viewpoints; as SCOUT explores the environment, semantically informed viewpoints are added. After ranking the viewpoints according to their semantic-geometric score (\S~\ref{subsec:viewpoint_scoring}), the robot moves to the highest-ranked viewpoint. Once the robot reaches that viewpoint, the captured color and depth images, together with the camera pose, are passed to the PSGG. The PSGG then processes these inputs to identify the objects visible in the current view (\S~\ref{subsec:frame_processing}). It next determines which objects are newly observed and which correspond to previously seen objects that should be merged with existing instances (\S~\ref{subsec:instantiation_and_fusion}). After this association step, the PSGG updates the uncertainty measure\footnote{In our current system, uncertainty is attached only to object hypotheses because UGT targets semantic ambiguity in object identity; relational edges are treated as structured context inferred from the current node geometry. In future work, we will attach uncertainty to relational edges as well.} associated with each object in the scene graph (\S~\ref{subsec:semantic_belief_update}, and the edge set(\S~\ref{subsec:edge_update}). These updates propagate back to the UGT, since uncertain nodes induce semantic viewpoints, and the system iterates until convergence. SCOUT declares convergence when both its geometric and semantic scores satisfy predefined thresholds. In our experiments, these thresholds were set empirically to 95\% geometric coverage and an uncertainty measure below 0.7 for all nodes. The following sections provide detailed descriptions of the UGT and PSGG modules.



\subsection{Uncertainty-guided Traversal (UGT)}
\label{sec:UGT}
Let $\mathcal{G}_t = (\mathcal{O}_t, \mathcal{E}_t)$ be the scene graph at time $t$, where $\mathcal{O}_t$ represents the set of nodes and $\mathcal{E}_t$ the set of edges. Each node $o_i \in \mathcal{O}_t$ stores fused 3D geometry, including a center and bounding box, a semantic belief $p_i^t(c)$ over the label vocabulary $\mathcal{L}$, the corresponding uncertainty measure $H_i^t$ (we use the Shannon entropy), and an observation history including past viewing bearings $\mathcal{H}_i$.
UGT consumes the current graph $\mathcal{G}_t$ together with an \textit{a priori} 2D occupancy map $\mathcal{M}$ from prior LiDAR scanning, with free and occupied cells $\mathcal{M}_{\mathrm{free}}$ and $\mathcal{M}_{\mathrm{occ}}$. Object geometry in $\mathcal{O}_t$ defines candidate view targets and supports visibility checks, node entropies $H_i^t$ prioritize semantically ambiguous objects, and histories $\mathcal{H}_i$ penalize redundant views. The map $\mathcal{M}$ faciliates the filtering of infeasible viewpoints and the estimation of traversal cost.

The goal of UGT is to generate the next set of valid candidate viewpoints, score them by semantic gain, geometric coverage gain, and travel cost, and greedily select the best one. This couples information gain with motion efficiency.
\subsubsection{Viewpoint Generation}
\label{subsec:viewpoint_generation}
To support both semantic exploitation and geometric exploration, we use a hybrid generator with a semantic set $\mathcal{V}_t^{\mathrm{sem}}$ for highly uncertain nodes and a geometric set $\mathcal{V}_t^{\mathrm{geo}}$ for systematic exploration. The final candidate set is
$\mathcal{V}_t = \mathcal{V}_t^{\mathrm{sem}} \cup \mathcal{V}_t^{\mathrm{geo}}$.

\paragraph{Semantic Coverage Candidates}
The semantic candidate set $\mathcal{V}_t^{\mathrm{sem}}$ targets nodes with high semantic uncertainty. For each of the top-$k$ most uncertain nodes $o_i$, we generate eight viewpoints uniformly on a circular ring centered at the node and orient them toward the node center. The ring radius scales with the 3D bounding-box diagonal to preserve a comparable view scale across object sizes, and candidates violating the free-space constraints of $\mathcal{M}$ are discarded. Thus, $\mathcal{V}_t^{\mathrm{sem}}$ concentrates samples in regions likely to yield large uncertainty reduction.

\paragraph{Geometric Coverage Candidates}
The geometric candidate set $\mathcal{V}_t^{\mathrm{geo}}$ targets unexplored free space. We uniformly sample reachable points in $\mathcal{M}_{\mathrm{free}}$ on a dense grid and generate eight headings per point over $[0, 2\pi)$ to encourage broad coverage without bias toward any single corridor or room orientation.

\subsubsection{Viewpoint Scoring}
\label{subsec:viewpoint_scoring}
For each candidate viewpoint $v \in \mathcal{V}_t$, we assign a weighted score that combines semantic gain and geometric coverage gain, scaled by travel cost:
$$
S_{\mathrm{final}}(v) = \frac{w_{\mathrm{cert}}\,S_{\mathrm{cert}}(v) + w_{\mathrm{cov}}\,S_{\mathrm{cov}}(v) }{S_{\mathrm{travel}}(v)}.
$$
Here, $S_{\mathrm{cert}}(v)$ is graph-based certainty gain, $S_{\mathrm{cov}}(v)$ is spatial coverage gain, and $S_{\mathrm{travel}}(v)$ is the shortest-path distance between the current robot pose $p_t$ and $v$, computed by $\text{A}^*$ \cite{hart1968formal} on $\mathcal{M}$. The weights satisfy $w_{\mathrm{cert}} + w_{\mathrm{cov}} = 1$.

We define $S_{\mathrm{cert}}(v)$ from two quantities: node visibility from viewpoint $v$ and observation novelty relative to past views.

\paragraph{Node Visibility}
We approximate the observability of a node from a candidate viewpoint by 3D ray casting. For viewpoint $v$, the image plane is uniformly subsampled to generate rays $\mathcal{R}(v)$ using the camera intrinsics and a fixed pixel stride. For each ray $r \in \mathcal{R}(v)$, we compute its first intersection with all node bounding boxes and retain only the nearest intersected node to account for occlusion.
We define the indicator function
$$
\mathbf{1}_{r \rightarrow o_i} =
\begin{cases}
1, & \text{if ray } r \text{ first intersects node } o_i, \\
0, & \text{otherwise,}
\end{cases}
$$
The visibility ratio of node $o_i$ from viewpoint $v$ is
\begin{equation}
\label{eq:visibility-score}
\mathrm{Vis}(v,i) =
\frac{1}{|\mathcal{R}(v)|}
\sum_{r \in \mathcal{R}(v)} \mathbf{1}_{r \rightarrow o_i},
\end{equation}
where $|\mathcal{R}(v)|$ is the number of sampled rays. This score is the fraction of rays for which node $o_i$ is the first visible surface.

\paragraph{View Novelty} 
We quantify viewpoint novelty relative to prior observations. For each node $o_i$, we maintain a set of historical viewing angles
$
\mathcal{H}_i = \{\theta_{i,k}\}_{k=1}^{K_i},
$
where $\theta_{i,k}$ is the $k$-th node-to-camera bearing. The minimum angular separation between the current viewpoint and the historical views is
$$
\Delta\theta(v,i) = \min_{k=1,\dots,K_i} \arccos\!\big(\cos(\theta(v,i) - \theta_{i,k})\big),
$$
where $\theta(v,i) = \operatorname{atan2}(y_v - y_i,\; x_v - x_i)$ denotes the node-to-camera bearing from viewpoint $v$ to node $o_i$, with $(x_v, y_v)$ the camera position and $(x_i, y_i)$ the node center in the $xy$-plane.

Let $\theta^\star \in (0,\pi]$ be the user-defined peak novelty angle (default $\frac{\pi}{4}$). The novelty score uses a piecewise remapped sine function:
\begin{equation} \label{eq:novelty-score}
\mathrm{Nov}(v,i) =
\begin{cases}
1, & \text{if } K_i = 0, \\ 
\sin\!\big(t(\Delta\theta(v,i))\big), & \text{otherwise},
\end{cases}
\end{equation}
where the remapping function $t(\Delta\theta)$ is given by
$$
t(\Delta\theta) =
\begin{cases}
\dfrac{\pi}{2}\dfrac{\Delta\theta}{\theta^\star}, & \Delta\theta \le \theta^\star, \\[8pt]
\dfrac{\pi}{2} + \dfrac{\pi}{2}\dfrac{\Delta\theta - \theta^\star}{\pi - \theta^\star}, & \Delta\theta > \theta^\star.
\end{cases}
$$

Thus, $\mathrm{Nov}(v,i)=0$ when the viewpoint repeats a previous bearing ($\Delta\theta=0$), reaches its maximum at $\Delta\theta=\theta^\star$, and decays back to $0$ as the viewpoint approaches the opposite direction ($\Delta\theta=\pi$). The remapping avoids over-rewarding nearly antipodal views, which often add little appearance evidence indoors.

\paragraph{Certainty-gain Score} 
Combining \textit{node visibility} and \textit{view novelty} that both lie in $[0,1]$, we define the \textit{observation strength} of viewpoint $v$ for node $o_i$ as
$$
\mathrm{Obs}(v,i) = \mathrm{Vis}(v,i)\,\mathrm{Nov}(v,i),
$$
which quantifies how strongly an RGB-D observation from $v$ can reduce the uncertainty of node $o_i$. We model this reduction with an exponential saturation function over the current uncertainty $H_i^t$:
$$
\Delta H(v,i) = H_i^t \left(1 - e^{-\alpha \,\mathrm{Obs}(v,i)}\right).
$$

The \textit{total certainty gain} of viewpoint $v$ is the sum over all nodes, or more efficiently over those visible from $v$:
$
G_{\mathrm{cert}}(v) = \sum_{o_i \in \mathcal{O}_t} \Delta H(v,i).
$
The \textit{certainty gain score} is the ratio between the certainty gain at viewpoint $v$ and the total remaining uncertainty in the scene graph:
\begin{equation}
S_{\mathrm{cert}}(v) =
\frac{G_{\mathrm{cert}}(v)}
{\sum_{o_i \in \mathcal{O}_t} H_i^t}.
\end{equation}

\paragraph{Coverage-gain Score} 
The coverage-gain score quantifies the expected increase in field-of-view (FOV) coverage over discretized space if an observation is taken from candidate viewpoint $v$. Let $\mathcal{T}$ be the set of target cells over which FOV coverage is evaluated, and let $\mathcal{C}_t \subseteq \mathcal{T}$ be the subset already covered by previous observations up to time $t$. We use $\mathrm{FOV}(v) \subseteq \mathcal{T}$ for the target cells that would fall within the geometric FOV of viewpoint $v$, and count only newly covered cells in $\mathrm{FOV}(v) \setminus \mathcal{C}_t$. Formally,
$$
S_{\mathrm{cov}}(v)=
\frac{\left|(\mathrm{FOV}(v)\cap \mathcal{T}) \setminus \mathcal{C}_t\right|}
{|\mathcal{T}|},
$$
which is the fraction of target cells newly covered by viewpoint $v$, normalized by $|\mathcal{T}|$. This term rewards views that expand the explored footprint even when no highly uncertain object is currently visible.



\subsection{Probabilistic Scene Graph Generator (PSGG)}
\label{sec:PSGG}
\begin{figure*}[!t]
\centering
\includegraphics[width=0.8\textwidth, height=6cm]{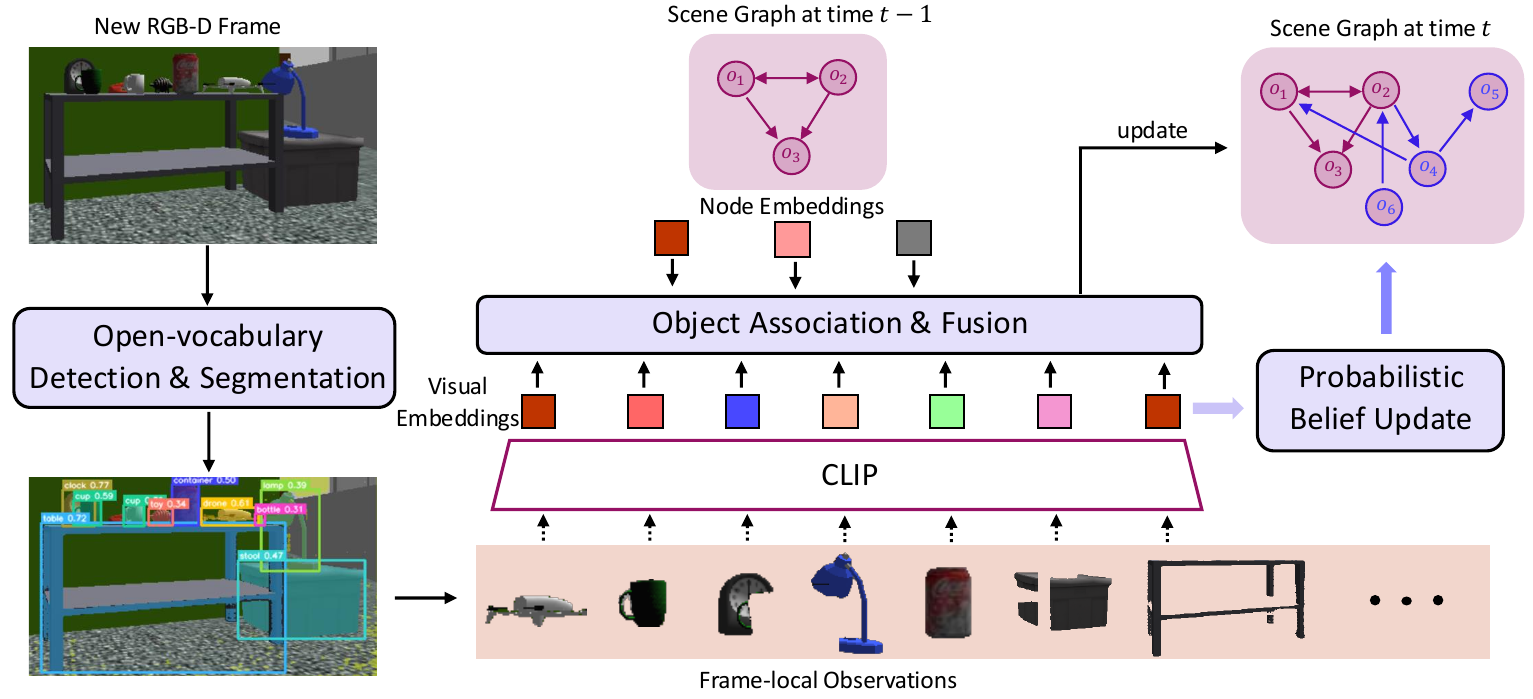}
\caption{Architecture of PSGG module.}
\label{fig:psgg-arch}
\end{figure*}

Given a new posed RGB-D observation $I_t$, PSGG consumes the last scene graph $\mathcal{G}_{t-1}$ and updates it to $\mathcal{G}_{t}$ incrementally by maintaining persistent object hypotheses and relational edges. It fuses 3D geometry, updates semantic beliefs and their entropies from frame-local evidence, records the observation history used by UGT, and refreshes structural relations in $\mathcal{E}_t$ (Fig.~\ref{fig:psgg-arch}).

\subsubsection{Per-frame Open-Vocabulary Observation Extraction}
\label{subsec:frame_processing}
For each incoming RGB-D frame $I_t$, we first integrate the depth observation into a fixed-bounds voxelized scene representation. The aligned RGB image is then processed by an open-vocabulary stack combining text-guided detection (Grounding DINO \cite{liu2024grounding}), mask generation (SAM \cite{kirillov2023segment}), and feature extraction (CLIP \cite{radford2021learning}). Given the class label from vocabulary $\mathcal{L}$, Grounding DINO proposes class-conditioned bounding boxes, which are filtered by confidence; each retained box prompts SAM to generate a segmentation mask, and the highest-scoring mask is kept as the 2D support points for the object observation; we then extract CLIP visual embedding $\mathbf{f}^{\mathrm{vis}}$ from both the original RGB (cropped to the bounding box) and its masked counterpart (cropped to the mask and padded with zeros) which are fused by means of a weighted sum, while the CLIP text embedding $\mathbf{f}^{\mathrm{text}}$ is obtained based on the associated class label. This process produces a set of frame-local object observations
$$
\mathcal{D}_t = \{d_j^t\}_{j=1}^{|\mathcal{D}_t|},
$$
where each observation $d_j^t$ contains a 2D mask, depth-backed lifted 3D support points (denoised by DBSCAN clustering), and extracted CLIP embedding descriptors $\mathbf{f}^{\mathrm{vis}}$ and $\mathbf{f}^{\mathrm{text}}$. PSGG treats these observations as \textit{noisy} and \textit{incomplete} evidence for updating persistent object hypotheses.

\subsubsection{Temporal Instance Association and 3D Fusion}
\label{subsec:instantiation_and_fusion}
PSGG associates each incoming observation $d_j^t$ with an existing object hypothesis by jointly evaluating spatial, visual, and textual similarities. For observation $d$ and object $o_i$, the association similarity is
$$
A(d,o_i) = \lambda_{\mathrm{sp}} S_{\mathrm{sp}}(d,o_i)
+ \lambda_{\mathrm{vis}} S_{\mathrm{vis}}(d,o_i)
+ \lambda_{\mathrm{text}} S_{\mathrm{text}}(d,o_i),
$$
where $S_{\mathrm{sp}}(d,o_i)$ measures geometric consistency between the 3D support of $d$ and the accumulated geometry of $o_i$ through the overlap ratio of their support points after a 3D bounding-box intersection check. $S_{\mathrm{vis}}(d,o_i)$ is the cosine similarity between the CLIP visual embedding $\mathbf{f}^{\mathrm{vis}}$ of $d$ and the aggregated CLIP visual embedding of $o_i$, and $S_{\mathrm{text}}(d,o_i)$ is the cosine similarity between the text embedding $\mathbf{f}^{\mathrm{text}}$ associated with $d$ and the aggregated text embedding of $o_i$. The weights satisfy $\lambda_{\mathrm{sp}} + \lambda_{\mathrm{vis}} + \lambda_{\mathrm{text}} = 1$. The highest association similarity score is thresholded: above the threshold, $d$ is merged into the matched node; otherwise, it initializes a new node (object hypothesis). Fused observations accumulate 3D support points, and in particular, the stored (aggregated) CLIP visual embedding is updated by a detection-count-weighted average between the previous node visual embedding and the new observation visual embedding $\mathbf{f}^{\mathrm{vis}}$, followed by $\ell_2$ normalization; the node text embedding is updated analogously.

\subsubsection{Probabilistic Semantic Belief Update}
\label{subsec:semantic_belief_update}
For each persistent object $o_i$, PSGG maintains a posterior distribution over the label vocabulary $\mathcal{L}$. Let $\mathcal{D}_i^t \subseteq \mathcal{D}_t$ denote the frame-$t$ observations associated with $o_i$. When $\mathcal{D}_i^t \neq \emptyset$, we average their visual embeddings $\mathbf{f}^{\mathrm{vis}}(d)$ to obtain an object-level frame feature
$
\mathbf{f}_i^t = \frac{1}{|\mathcal{D}_i^t|}\sum_{d \in \mathcal{D}_i^t} \mathbf{f}^{\mathrm{vis}}(d).
$
For each class label $c \in \mathcal{L}$, let $\mathbf{e}_c$ denote its CLIP text embedding. We define the \textit{in-frame belief} as the soft-maxed cosine similarity
$$
q_i^t(c) = \frac{\exp\left(\tau\,\mathrm{sim}(\mathbf{f}_i^t,\mathbf{e}_c)\right)}
{\sum_{c' \in \mathcal{L}} \exp\left(\tau\,\mathrm{sim}(\mathbf{f}_i^t,\mathbf{e}_{c'})\right)},
$$
where $\mathrm{sim}(\cdot,\cdot)$ is cosine similarity and $\tau$ is a temperature parameter. With $p_i^{t-1}(c)$ the prior before processing frame $I_t$, the posterior belief is updated by multiplicative Bayesian fusion:
$$
p_i^t(c)=\frac{p_i^{t-1}(c)\,q_i^t(c)}{\sum_{c'\in\mathcal{L}} p_i^{t-1}(c')\,q_i^t(c')}.
$$
The uncertainty of node $o_i$ is then computed as the Shannon entropy
$
H_i^t = -\sum_{c \in \mathcal{L}} p_i^t(c)\log p_i^t(c).
$
This preserves ambiguity across similar labels instead of forcing an early hard decision. 

\subsubsection{Incremental Scene Graph Construction}
\label{subsec:edge_update}
After updating node attributes in $\mathcal{O}_t$ (\eg{} visual and text embeddings, semantic beliefs, and 3D geometry), PSGG updates the relational edge set $\mathcal{E}_t$ to capture scene structure. Each edge type has a specific meaning: a directed \textit{inside} edge indicates containment; a directed \textit{on} edge indicates physical support; a directed \textit{belong} edge denotes a part-to-whole relation, such as a handle belonging to a cabinet; and an undirected \textit{next to} edge denotes local adjacency when no stronger structural relation is present. Such a construction is based on the geometric classification of the updated 3D bounding boxes. 


\section{Numerical Evaluation}
\begin{figure}[htbp]
    \centering
    \begin{subfigure}[b]{0.255\textwidth}
        \centering
        \includegraphics[width=\textwidth]{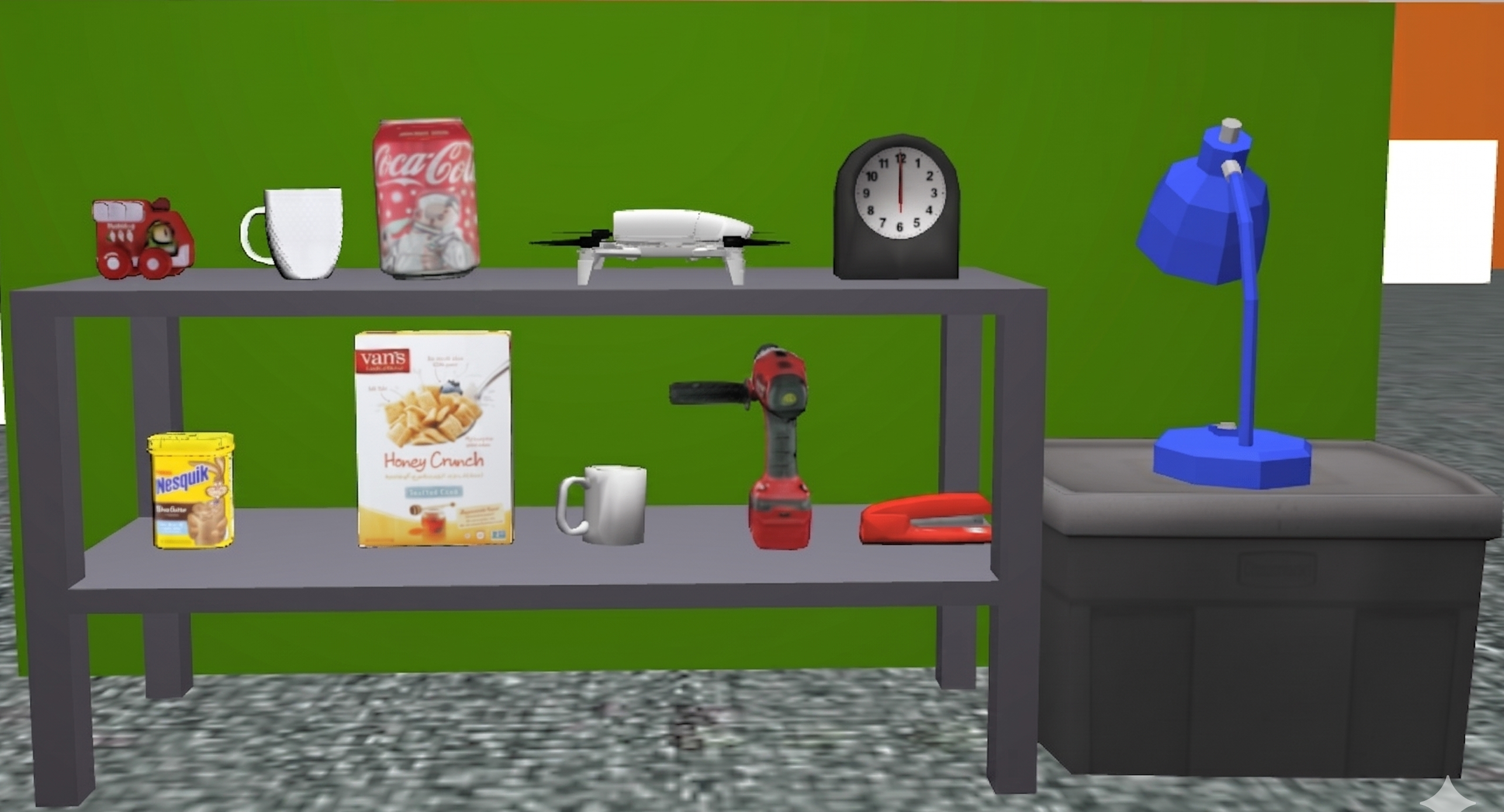}
        \caption{Simple scenario}
        \label{fig:simple_scenario}
    \end{subfigure}
    \hfill
    \begin{subfigure}[b]{0.22\textwidth}
        \centering
        \includegraphics[width=\textwidth]{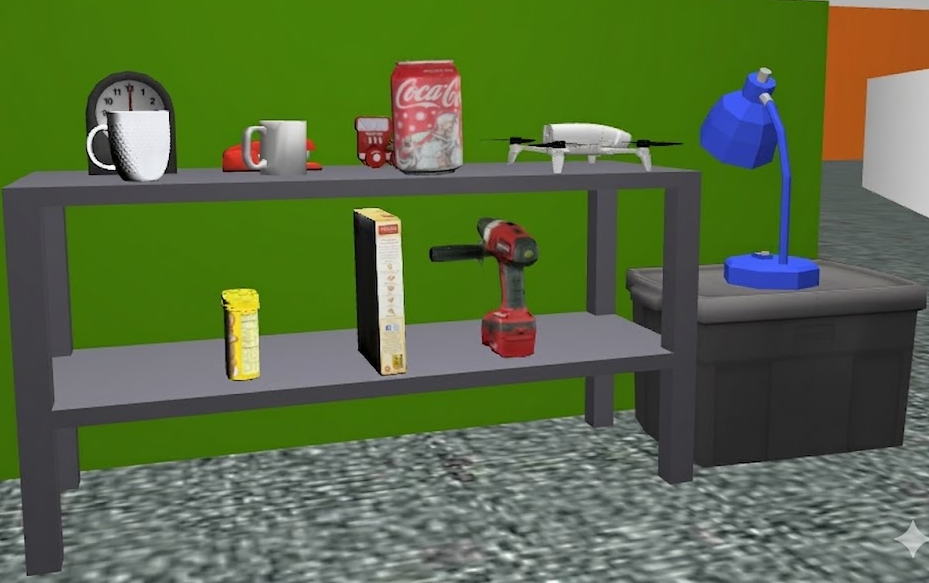}
        \caption{Challenging scenario}
        \label{fig:challenging_scenario}
    \end{subfigure}
    
    \caption{The two scenarios used for numerical evaluation.}
    \label{fig:evaluation_scenarios}
\end{figure}

In this section, we present preliminary results for SCOUT. We evaluate the system in simulation using Gazebo~\cite{koenig2004design} together with a hand-crafted, high-fidelity model designed to capture the mechanical behavior of real robots and their operating environment. To assess performance, we construct two simulated environments, denoted as \emph{simple} and \emph{challenging}, shown in Fig.~\ref{fig:evaluation_scenarios}.

Both environments share the same overall layout: a two-level shelf containing 10 objects and a container holding 1 object, for a total of 13 objects in the scene. In the simple scenario, all objects are front-facing and arranged side by side, resulting in no occlusions (Fig.~\ref{fig:simple_scenario}). In the challenging scenario (Fig.~\ref{fig:challenging_scenario}), two objects, a cereal box and a canister, are rotated by $90^\circ$, and three objects are partially or heavily occluded: a clock behind a mug with only its top visible, a stapler behind a mug with its sides exposed, and a small toy behind a can that is almost entirely occluded.

We compare SCOUT against a lawnmower traversal baseline designed to represent a decoupled scene-graph generation pipeline, as is common in prior work. In this baseline, data collection is separated from scene-graph generation: the robot first executes a fixed set of viewpoints generated offline from a prior 2D map to geometrically sweep the area of interest, and the scene graph is then constructed from the collected observations. Consequently, viewpoint selection is not influenced by the evolving scene graph or by semantic uncertainty during execution. We report node- and edge-level precision and recall for both methods. Precision measures the fraction of detected nodes or edges that are correctly classified, while recall measures the fraction of ground-truth nodes or edges that are correctly identified. In the simple scenario, the ground-truth edge set consists of 11 \textit{on} relationships, corresponding to the ten objects on the shelf and the lamp on the container, and 11 \textit{next-to} relationships. Two objects are considered \textit{next to} each other if no third object lies between them. In the challenging scenario, the ground-truth edge set contains 11 \textit{on} relationships and 18 \textit{next-to} relationships. For each method-scenario pair, we perform three trials and report average performance. For SCOUT, we cap the maximum number of observations at 50. The results are summarized in Table~\ref{tab:evaluation_results}.

\begin{table}[h]
\centering
\caption{Node-level and edge-level precision and recall for SCOUT and lawnmower.}
\label{tab:evaluation_results}
\begin{tabular}{llcccc}
\hline
\multirow{2}{*}{\textbf{Scenario}} & \multirow{2}{*}{\textbf{Method}} & \multicolumn{2}{c}{\textbf{Nodes}} & \multicolumn{2}{c}{\textbf{Edges}} \\
 & & \textbf{Precision} & \textbf{Recall} & \textbf{Precision} & \textbf{Recall} \\ \hline
\multirow{2}{*}{Simple}
  & SCOUT     & 1.00 & 1.00 & 1.00 & 0.62 \\
  & Lawnmower & 0.86 & 0.79 & 0.86 & 0.48 \\ \hline
\multirow{2}{*}{Challenging}
  & SCOUT     & 0.93 & 0.94 & 0.67 & 0.39 \\
  & Lawnmower & 0.72 & 0.74 & 0.54 & 0.27 \\ \hline
\end{tabular}
\end{table}

The results show that SCOUT outperforms the lawnmower traversal baseline in both node- and edge-level precision and recall. In the simple scenario, SCOUT achieves perfect node detection performance, with 100\% precision and 100\% recall. In the challenging scenario, SCOUT maintains strong node-level performance, achieving 93\% precision and 94\% recall, compared with 72\% precision and 74\% recall for the lawnmower baseline.

\begin{figure}[htbp]
    \centering
    \begin{subfigure}[b]{0.5\textwidth}
        \centering
        \includegraphics[width=\textwidth]{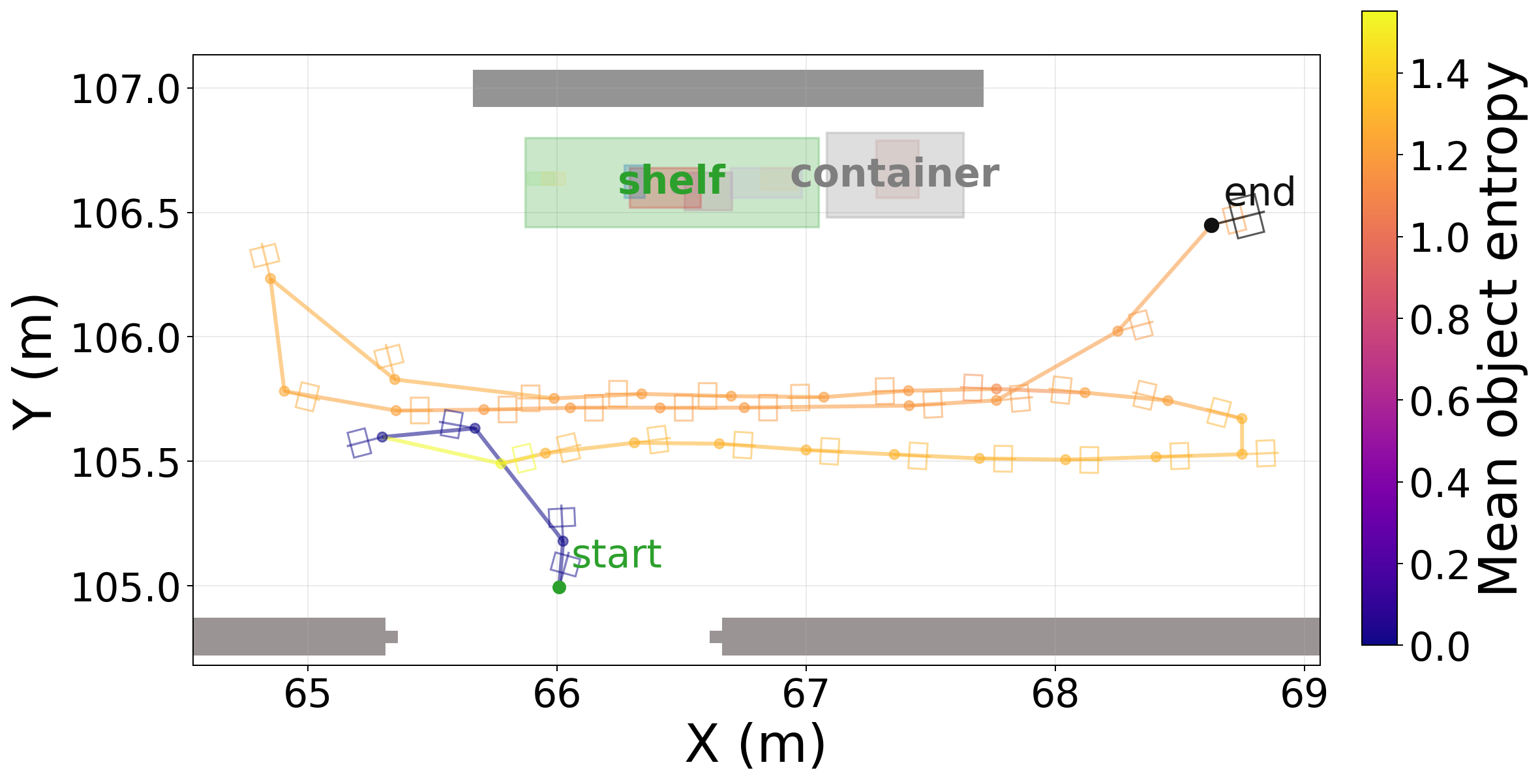}
        \caption{Lawnmower}
        \label{fig:lawnmower_cam_traj}
    \end{subfigure}
    \hfill
    \begin{subfigure}[b]{0.5\textwidth}
        \centering
        \includegraphics[width=\textwidth]{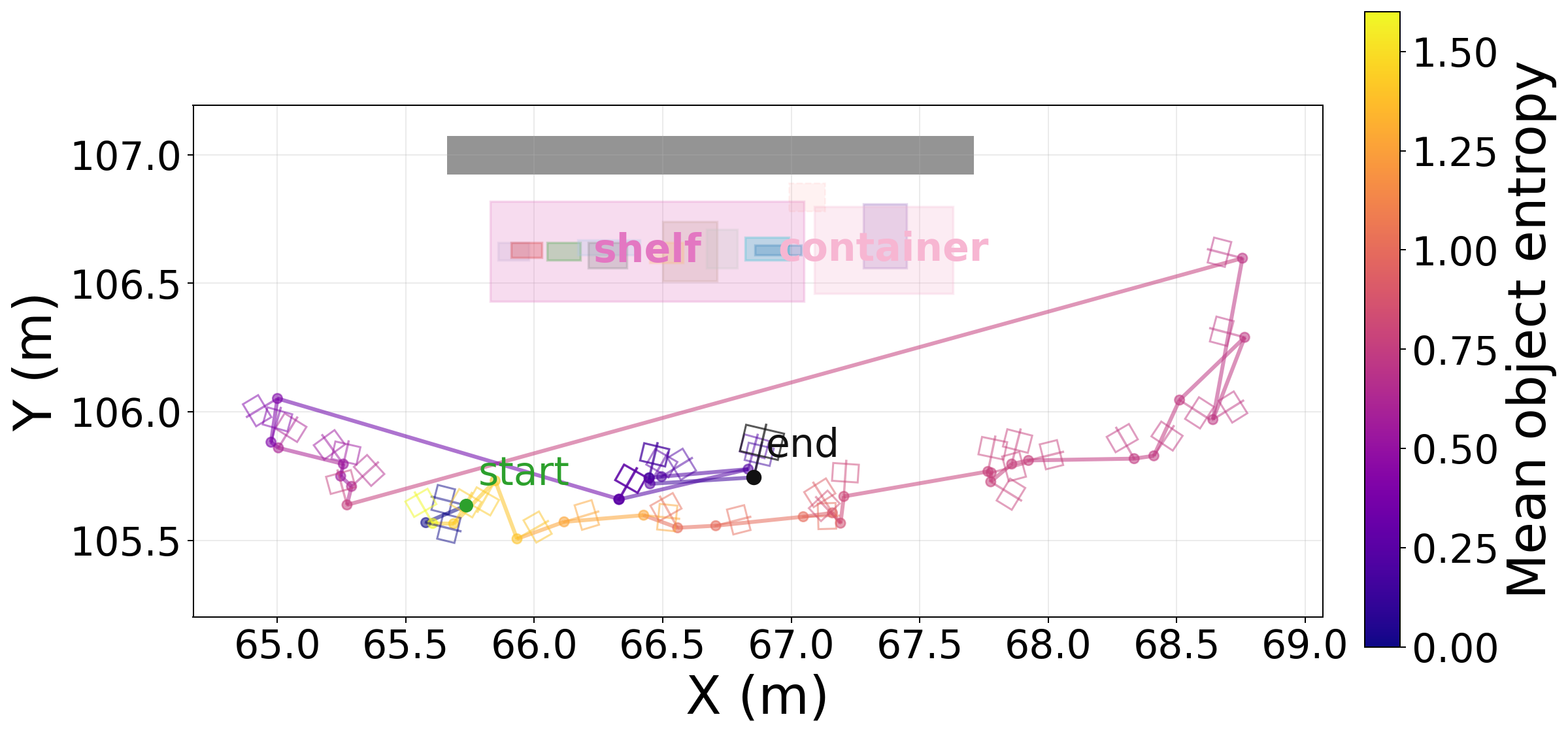}
        \caption{SCOUT}
        \label{fig:scout_cam_traj}
    \end{subfigure}
    
    \caption{Camera trajectory for the lawnmower baseline and SCOUT in the simple scenario.}
    \label{fig:simple_scenario_trajectories}
\end{figure}

\begin{figure}[t]
    \centering
    \includegraphics[width=\linewidth]{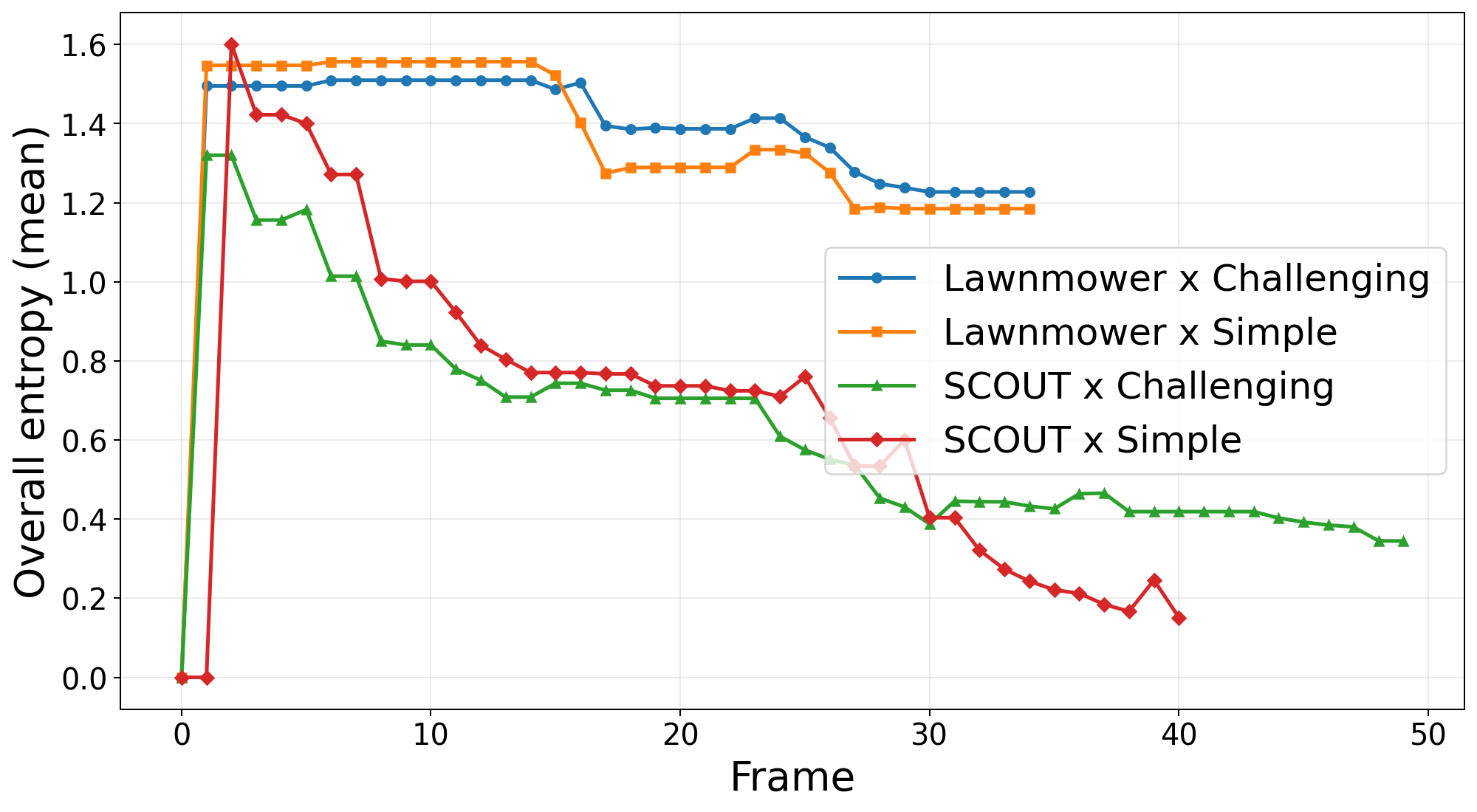}
    \caption{Mean entropy over time for SCOUT and lawnmower.}
    \label{fig:entropy_over_time}
\end{figure}

These quantitative results are consistent with the qualitative behaviors shown in Fig.~\ref{fig:lawnmower_cam_traj} and~\ref{fig:scout_cam_traj}. In the simple scenario, the lawnmower baseline follows a geometric sweep of the environment, whereas SCOUT explicitly revisits semantically informative regions to refine its scene understanding. This behavior produces a more accurate scene representation and reduces overall uncertainty. The effect is also reflected in Fig.~\ref{fig:entropy_over_time}, where SCOUT exhibits a steady reduction in mean entropy as new observations are incorporated, while the lawnmower baseline shows a more stepwise decrease that occurs primarily when semantically rich regions are re-observed during the sweep.

It is worth noting that the mean entropy for SCOUT occasionally plateaus, for example between approximately frames 15 and 25. This behavior arises from SCOUT’s design, which balances semantic exploration with geometric coverage. In these scenarios, SCOUT uses this phase to explore the environment and verify whether additional regions of interest exist. In the simple scenario, entropy resumes a clear downward trend after this plateau, indicating that SCOUT has identified the relevant region and subsequently concentrates its observations there.

A second plateau appears in the challenging scenario around frame 30. At this stage, SCOUT has largely resolved the scene structure, and the remaining uncertainty is concentrated in a small number of difficult cases, specifically associating the two visible sides of the rotated cereal box as a single object and acquiring informative viewpoints of the partially occluded stapler. The latter is particularly challenging because pose and orientation tolerances in the navigation stack prevent the robot from consistently reaching the exact requested viewpoint, requiring multiple attempts before a sufficiently informative observation of the stapler can be obtained.

At the edge level, SCOUT successfully detects all \textit{on} relationships in the simple scenario; however, it fails to recover many of the \textit{next-to} relationships, resulting in an overall edge recall of 62\%. We expect that further system tuning, particularly of the proximity threshold, would improve this result. In the challenging scenario, edge-level performance degrades further across both relation types. These results are significantly affected by an object association failure in one run, where the system did not merge two views of the same shelf before the allotted frame limit was reached. This error produced a duplicate shelf entity and a large number of associated \textit{on} relationships, which dominated the false-positive count.

Finally, in the challenging scenario, our system currently does not converge before reaching the prescribed frame limit. This occurs because the maximum entropy is not reduced below the specified threshold within the available observation budget. Improving the trade-off between semantic certainty and convergence time remains an important direction for future work.

\section{Conclusion}

SCOUT couples active traversal with scene graph construction to enable uncertainty-guided exploration of physical environments. By leveraging a 2D map prior and explicitly balancing geometric gain with semantic uncertainty reduction, SCOUT identifies semantically informative regions and revisits them to refine its understanding of the scene. Our preliminary results show that SCOUT yields clear improvements over a lawnmower baseline, highlighting the value of integrating semantic reasoning into the traversal process. Future work will focus on comprehensive evaluations, improving the efficiency and convergence of the system, and running validations in real-world robotic deployments.

\bibliographystyle{IEEEtran}
\bibliography{ref}

@inproceedings{gu2024conceptgraphs,
  title={Conceptgraphs: Open-vocabulary 3d scene graphs for perception and planning},
  author={Gu, Qiao and Kuwajerwala, Ali and Morin, Sacha and Jatavallabhula, Krishna Murthy and Sen, Bipasha and Agarwal, Aditya and Rivera, Corban and Paul, William and Ellis, Kirsty and Chellappa, Rama and others},
  booktitle={2024 IEEE International Conference on Robotics and Automation (ICRA)},
  pages={5021--5028},
  year={2024},
  organization={IEEE}
}

@inproceedings{werby2024hierarchical,
  title={Hierarchical open-vocabulary 3d scene graphs for language-grounded robot navigation},
  author={Werby, Abdelrhman and Huang, Chenguang and B{\"u}chner, Martin and Valada, Abhinav and Burgard, Wolfram},
  booktitle={First Workshop on Vision-Language Models for Navigation and Manipulation at ICRA 2024},
  year={2024}
}

@article{linok2024beyond,
  title={Beyond bare queries: Open-vocabulary object retrieval with 3d scene graph},
  author={Linok, Sergey and Zemskova, Tatiana and Ladanova, Svetlana and Titkov, Roman and Yudin, Dmitry A},
  journal={CoRR},
  year={2024}
}

@article{hughes2022hydra,
  title={Hydra: A real-time spatial perception system for 3D scene graph construction and optimization},
  author={Hughes, Nathan and Chang, Yun and Carlone, Luca},
  journal={arXiv preprint arXiv:2201.13360},
  year={2022}
}

@inproceedings{armeni20193d,
  title={3d scene graph: A structure for unified semantics, 3d space, and camera},
  author={Armeni, Iro and He, Zhi-Yang and Gwak, JunYoung and Zamir, Amir R and Fischer, Martin and Malik, Jitendra and Savarese, Silvio},
  booktitle={Proceedings of the IEEE/CVF international conference on computer vision},
  pages={5664--5673},
  year={2019}
}

@article{hart1968formal,
  title={A formal basis for the heuristic determination of minimum cost paths},
  author={Hart, Peter E and Nilsson, Nils J and Raphael, Bertram},
  journal={IEEE transactions on Systems Science and Cybernetics},
  volume={4},
  number={2},
  pages={100--107},
  year={1968},
  publisher={IEEE}
}

@inproceedings{liu2024grounding,
  title={Grounding dino: Marrying dino with grounded pre-training for open-set object detection},
  author={Liu, Shilong and Zeng, Zhaoyang and Ren, Tianhe and Li, Feng and Zhang, Hao and Yang, Jie and Jiang, Qing and Li, Chunyuan and Yang, Jianwei and Su, Hang and others},
  booktitle={European conference on computer vision},
  pages={38--55},
  year={2024},
  organization={Springer}
}

@inproceedings{kirillov2023segment,
  title={Segment Anything},
  author={Kirillov, Alexander and Mintun, Eric and Ravi, Nikhila and Mao, Hanzi and Rolland, Chloe and Gustafson, Laura and Xiao, Tete and Whitehead, Spencer and Berg, Alexander C. and Lo, Wan-Yen and Dollar, Piotr and Girshick, Ross},
  booktitle={Proceedings of the IEEE/CVF International Conference on Computer Vision},
  pages={4015--4026},
  year={2023},
  doi={10.1109/ICCV51070.2023.00371},
}

@inproceedings{radford2021learning,
  title={Learning Transferable Visual Models From Natural Language Supervision},
  author={Radford, Alec and Kim, Jong Wook and Hallacy, Chris and Ramesh, Aditya and Goh, Gabriel and Agarwal, Sandhini and Sastry, Girish and Askell, Amanda and Mishkin, Pamela and Clark, Jack and Krueger, Gretchen and Sutskever, Ilya},
  booktitle={Proceedings of the 38th International Conference on Machine Learning},
  pages={8748--8763},
  year={2021},
  volume={139},
  series={Proceedings of Machine Learning Research},
  publisher={PMLR},
}

@inproceedings{koenig2004design,
  title={Design and use paradigms for gazebo, an open-source multi-robot simulator},
  author={Koenig, Nathan and Howard, Andrew},
  booktitle={2004 IEEE/RSJ international conference on intelligent robots and systems (IROS)(IEEE Cat. No. 04CH37566)},
  volume={3},
  pages={2149--2154},
  year={2004},
  organization={Ieee}
}

@article{settles2009active,
  title={Active Learning Literature Survey},
  author={Settles, Burr},
  journal={Computer Sciences Technical Report 1648, University of Wisconsin--Madison},
  year={2009}
}

\end{document}